\documentclass[sigconf]{acmart}
\pdfoutput=1
\makeatletter
\def\@ACM@checkaffil{
    \if@ACM@instpresent\else
    \ClassWarningNoLine{\@classname}{No institution present for an affiliation}%
    \fi
    \if@ACM@citypresent\else
    \ClassWarningNoLine{\@classname}{No city present for an affiliation}%
    \fi
    \if@ACM@countrypresent\else
        \ClassWarningNoLine{\@classname}{No country present for an affiliation}%
    \fi
}
\makeatother

\AtBeginDocument{%
  \providecommand\BibTeX{{%
    \normalfont B\kern-0.5em{\scshape i\kern-0.25em b}\kern-0.8em\TeX}}}

\copyrightyear{2024}
\acmYear{2024}
\setcopyright{acmlicensed}
\acmConference[KDD '24] {Proceedings of the 30th ACM SIGKDD Conference on Knowledge Discovery and Data Mining }{August 25--29, 2024}{Barcelona, Spain.}
\acmBooktitle{Proceedings of the 30th ACM SIGKDD Conference on Knowledge Discovery and Data Mining (KDD '24), August 25--29, 2024, Barcelona, Spain}
\acmISBN{979-8-4007-0490-1/24/08}
\acmDOI{10.1145/3637528.3671773}

\usepackage{multirow}
\usepackage{bm}
\usepackage{textcomp}
\usepackage{subfigure}
\usepackage[normalem]{ulem}
\useunder{\uline}{\ul}{}
\usepackage{stfloats}
\usepackage{pifont}
\usepackage{svg} 
\usepackage{footnote} 
\usepackage[bottom]{footmisc}
\begin{document}
\title{DyGKT: Dynamic Graph Learning for Knowledge Tracing}


\author{Ke Cheng}
\email{ckpassenger@buaa.edu.cn}
\authornote{These authors contributed equally to this work.}
\affiliation{%
  \institution{SKLSDE Lab, Beihang University}
  \city{Beijing}
  \country{China}
}

\author{Linzhi Peng}
\email{lzpeng626@buaa.edu.cn}
\authornotemark[1]
\affiliation{%
  \institution{SKLSDE Lab, Beihang University}
  \city{Beijing}
  \country{China}
}

\author{Pengyang Wang}
\email{pywang@um.edu.mo}
\affiliation{%
  \institution{SKL-IOTSC, Department of Computer and Information Science, University of Macau}
  \city{Macau}
  \country{China}
}

\author{Junchen Ye}
\email{junchenye@buaa.edu.cn}
\authornote{Corresponding Author.}
\affiliation{%
  \institution{School of Transportation Science and Engineering, Beihang University}
  \city{Beijing}
  \country{China}
}

\author{Leilei Sun}
\email{leileisun@buaa.edu.cn}
\affiliation{%
  \institution{SKLSDE Lab, Beihang University}
  \city{Beijing}
  \country{China}
}

\author{Bowen Du}
\email{dubowen@buaa.edu.cn}
\affiliation{%
  \institution{Zhongguancun Laboratory}
}
\affiliation{%
  \institution{School of Transportation Science and Engineering, Beihang University}
  \city{Beijing}
  \country{China}
}

\renewcommand{\shortauthors}{Ke Cheng et al.}

\begin{abstract}
Knowledge Tracing aims to assess student learning states by predicting their performance in answering questions. Different from the existing research which utilizes fixed-length learning sequence to obtain the student states and regards KT as a static problem, this work is motivated by three dynamical characteristics: 1) The scales of students answering records are constantly growing; 2) The semantics of time intervals between the records vary; 3) The relationships between students, questions and concepts are evolving. 
The three dynamical characteristics above contain the great potential to revolutionize the existing knowledge tracing methods.
Along this line, we propose a Dynamic Graph-based Knowledge Tracing model, namely DyGKT.
In particular, a continuous-time dynamic question-answering graph for knowledge tracing is constructed to deal with the infinitely growing answering behaviors, and it is worth mentioning that it is the first time dynamic graph learning technology is used in this field.
Then, a dual time encoder is proposed to capture long-term and short-term semantics among the different time intervals. 
Finally, a multiset indicator is utilized to model the evolving relationships between students, questions, and concepts via the graph structural feature.
Numerous experiments are conducted on five real-world datasets, and the results demonstrate the superiority of our model. All the used resources are publicly available at https://github.com/PengLinzhi/DyGKT.

\end{abstract}

\ccsdesc[500]{Computing methodologies~Artificial intelligence}
\ccsdesc[500]{Information systems~Data mining}


\keywords{Educational Data Mining, Knowledge Tracing, Graph Neural Networks, Dynamic Graph}

\maketitle

\section{Introduction}
\label{section-1}
\begin{figure}[t]
\includegraphics[width=1\linewidth]{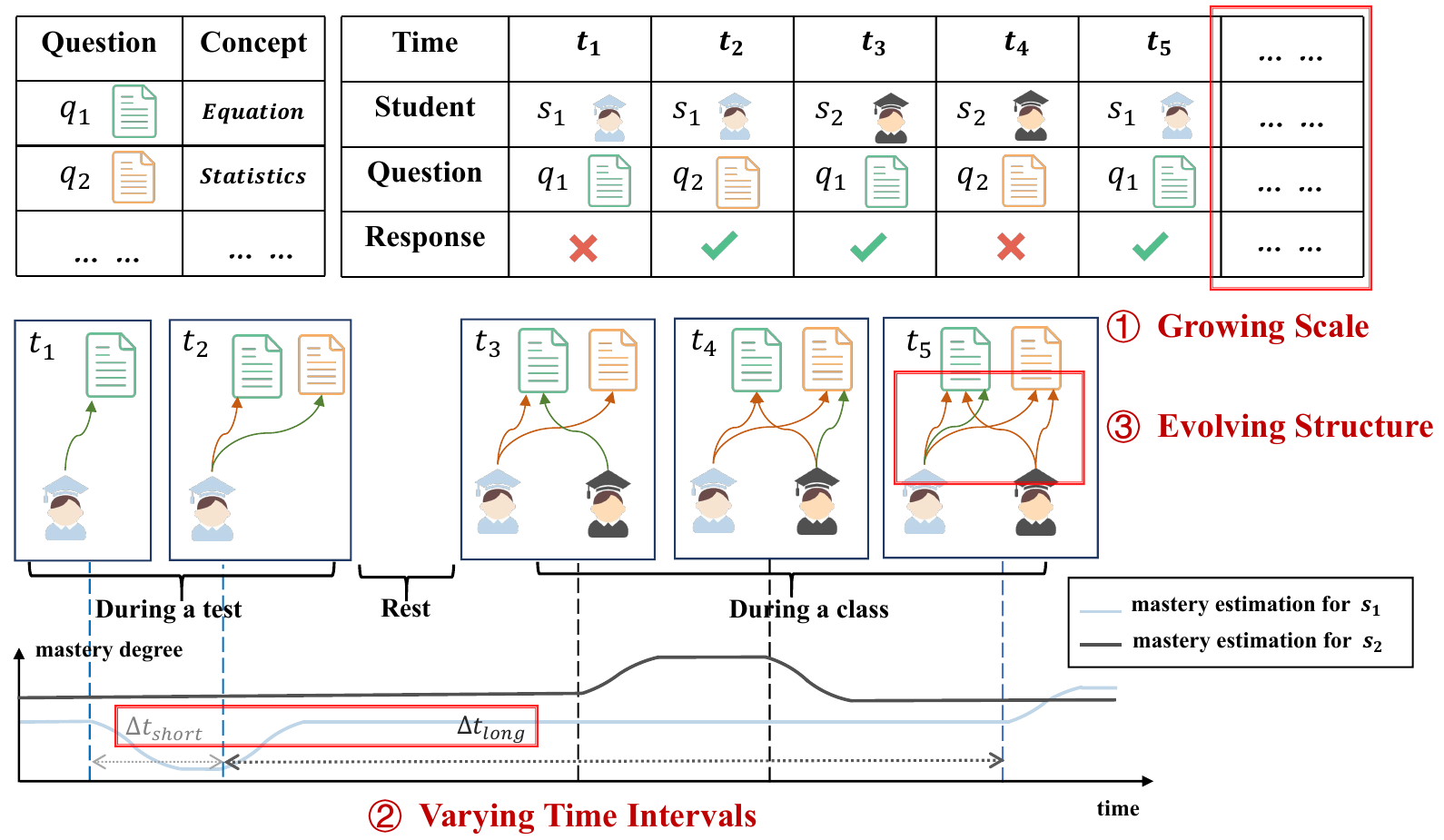}
\vspace{-0.3cm}
    \caption{Knowledge tracing with dynamic graph to trace students' learning states. This task has three dynamics: 1) constantly growing scale; 2) varying semantics of time intervals between long and short time intervals; 3) evolving relationships between students, questions, and concepts.}
\vspace{-0.4cm}
\label{Fig:progress}
\end{figure}

Nowadays, online education platforms are rapidly developing and have accumulated a large amount of educational records. Among intelligent tutoring systems, knowledge tracing (KT) models leverage students' responses to a set of questions to track their evolving knowledge states over time. The results obtained from knowledge tracing can be widely applied in downstream tasks such as predicting students' future performance and recommending learning materials.
Traditional knowledge tracing models initially relied on Bayesian and Factor Analysis. 
Recently, deep learning Knowledge Tracing (DLKT) has been proposed by applying deep learning techniques to the KT task, which can better capture a student's temporal dynamic knowledge state. DLKT models first gather a fix-lengthed question-answer record sequence for each student and then predict the student's performance on the next question based on those static sequences. DKT \cite{piech2015deep} utilizes RNN \cite{lipton2015critical} or LSTM \cite{hochreiter1997long} to trace the student learning process. AKT\cite{ghosh2020context} uses self-attention networks for capturing temporal dependencies in knowledge tracing. 
Based on the performance sequence, some works augment student performance modeling by enhancing the model with additional features.
For example, time interval is considered in forgetting behavior in current KT methods. DKT-Forgetting \cite{nagatani2019augmenting} and RKT \cite{pandey2020rkt} provide a linear or static exponential decay forgetting model for time intervals in the question-answering process.
CT-NCM\cite{ma2022reconciling} utilized the Hawkes Process to model the temporal effect of different time intervals in the sequence and use LSMT to predict student performance. Other models extract the relationship between concepts or questions for the structure information in the KT graph\cite{nakagawa2019graph,yang2021gikt}.

However, we argue that current works ignore three dynamic patterns in the KT task:

Firstly, the \textbf{scale} pattern acknowledges that \textbf{the number of students answering records is constantly growing}. Existing KT models only evaluate by single-step predicting for each student on static sequences, and utilize ID-embedding to represent the knowledge state of the student or the difficulty of the question. but in real-world scenarios, the length of learning records is continuously increasing, the prediction should be multi-step and robust on unseen questions. Sequence-based problem formalization on KT can not fit the real-world application, and the ID-embedding-based representation does not provide information for questions not contained in the training data.

Secondly, the \textbf{temporal} pattern refers to the fact that \textbf{the meaning of long and short time intervals varies}. A student's learning record is an irregular time series that changes dynamically, containing both long and short time intervals. For instance, a student might answer many questions in one test and then take long breaks before engaging in the next class or test. Short intervals reflect a student's familiarity with the question, while long intervals indicate forgetting behavior. Existing static and decay time modules that model forgetting behavior cannot extract the temporal dynamics in diverse scenarios.

Thirdly, the \textbf{structural} pattern involves \textbf{the complex relationship between students, questions, and concepts}. A question can be solved by multiple students before a certain moment, or be repeatedly practiced by one student. Therefore, there are numerous links between a student and a question, forming link multisets.
For example, if a student repeats guessing a question and finally comes up with the correct answer through traversal, it is obvious that the student does not learn by the question, models that cannot distinguish between repetitive behaviors may be confused by these abnormal student performances.

In this paper, we observe that the continuous-time dynamic graph (CTDG) is a suitable framework for depicting the three dynamic patterns by representing entities as nodes and capturing their dynamic interaction links with continuous timestamps. We provide a visualization case of knowledge tracing using CTDG in Figure \ref{Fig:progress}. The CTDG framework is particularly useful for representing evolving nodes as the interaction happens.
However, existing CTDG methods utilize Fourier Time encoding to capture the temporal effect which can hardly model the temporal pattern in the learning process. Also, 
CTDG methods are mostly designed for bipartite graphs, treating the student answering a question as a single link, which can not capture the complex structural pattern in KT either.

To address the issue above, we propose a novel method
named called \textbf{Dy}namic \textbf{G}raph-based \textbf{K}nowledge \textbf{T}racing (\textbf{DyGKT}).
This method aims to trace students’ dynamic learning process through a continuous time dynamic graph. In this graph, students and questions are represented as nodes, while their continuous interactions are represented as temporal links. 
Our goal is to classify the label of the link between two nodes at a specific time, which represents the student's performance on the given problem at that time. This newly designed task can adapt to infinitely growing data dynamically. To address temporal dynamics, we design a dual time encoding strategy to differentiate between long-term rest and short-term thinking patterns. For structural dynamics, we propose a multiset indicator to label elements that belong to the link multiset of the student, the question, and the concept related to the question-answer performance that is about to be predicted.
We conducted extensive experiments on five datasets, and the results demonstrate the effectiveness of the proposed method.

Our contributions can be summarized as follows:
\begin{itemize}
\item A continuous dynamic graph-based model for knowledge tracing problems is proposed. Different from the existing work which extracts fixed-length question-answering sequences, we extend it to the infinite-length scale, making it more suitable for real-world scenarios.
\item A dual-time encoding strategy is proposed to distinguish between the long-term and short-term semantics by a threshold. Compared with the existing work which utilizes a single decay or static function, the simple yet effective method we proposed could better describe the student behaviors between different time intervals.

\item We bring out a multiset indicator to deal with the complex evolving relationship. We extend existing CTDG methods that treat question-answer as a single link and utilize this module to construct the evolving structural features based on the repeat behavior by picking out the multiple links between students, questions, and corresponding concepts.


\end{itemize}

\section{Related work}
\label{section-5}
\subsection{Knowledge Tracing}

With the development of deep learning, deep learning techniques are introduced into the KT task, forming Deep Learning Knowledge Tracing(DLKT) which has achieved promising performance. DLKT methods aim to capture the temporal dynamics in a sequence of interactions \cite{lipton2015critical,sutskever2014sequence,hochreiter1997long}. 
DKT \cite{piech2015deep} and other extended DKT models take a student's previous question answers as input and use RNNs or LSTMs to represent the student's state by incorporating additional features such as answering rates and time spent on practicing \cite{xiong2016going}, adding a question's ID\cite{sonkar2020qdkt} or known difficulty\cite{shen2022assessing} as the question's extensive feature, or augmenting loss function\cite{yeung2018addressing, liu2023enhancing}.
Some KT models are evolving beyond temporal sequence models,
Attentive Knowledge Tracing Models incorporate an attention mechanism, allowing the model to assign importance weights to questions based on their interactions, capturing their relative significance.SAKT, AKT, SAINT and simpleKT use the self-attention network to capture the relevance between the KCs and the students’ historical interactions \cite{pandey2019self,ghosh2020context,liu2023simplekt,choi2020towards}. 

The effectiveness of learning is related to time. Examples include forgetting after a long gap and practicing several questions within one class.
Some models consider forgetting as a process that leads to a decline in learning states over time, 
DKT-t\cite{lalwani2019does} and DKT-Forgetting\cite{nagatani2019augmenting} incorporate decline functions over time into DKT. LPKT\cite{shen2021learning} considers the time-related effects of both learning and forgetting behaviors simultaneously.
Others believe that the forgetting rate for each individual and concept is unique, for instance, 
forgetting models based on Hawkes theory assign different forgetting rates to different concepts\cite{wang2021temporal,ma2022reconciling,jiang2021eduhawkes}. 

Another significant factor is the spatial feature of questions and concepts. 
Memory Augmented KT models statically\cite{miller2016key} or dynamically\cite{zhang2017dynamic} store the relationships between concepts and questions by external key-value memory space to keep longer sequence dependencies.
GKT\cite{nakagawa2019graph}, GIKT\cite{yang2021gikt}, HIN\cite{jiang2023multi} utilize graph neural networks to create concepts and question relationship graphs.

\subsection{Dynamic Graph Learning}

Dynamic graphs denote entities as nodes and represent their interactions as edges with timestamps. This is a powerful way to model the dynamics and evolution of complex systems over time.
In particular, continuous-time dynamic Graphs (CTDG) \cite{trivedi2017knowevolve} are represented as timed lists of node interaction events.

CTDG methods efficiently manage computation costs by partitioning event sequences into fixed-length edge batches and transmitting information within subgraphs. 
Key aspects of network evolution include temporal and structural information. For example, 
Jodie\cite{kumar2019predicting} models the user’s evolution with an RNN-based encoder when interaction occurs. 
TGAT\cite{xu2020inductive} extracts and encodes a local temporal subgraph to represent a node’s current state. TGN\cite{DBLP:journals/corr/abs-2006-10637} extends Jodie with a graph convolutional layer to aggregate information from local neighbors and update node representation with the aggregated temporal information.

In the downstream tasks of dynamic graph learning, applications range from recommendation systems in social networks\cite{alvarez2021evolutionary, song2019session, kumar2019predicting}, financial networks\cite{ranshous2015anomaly, wang2021bipartite, chang2021f}, and user-item interaction systems\cite{fan2021continuous, yu2022element, yu2022modelling}.
Continuous exploration and validation of CTDG methods in various contexts are crucial for unlocking their full potential.

\section{Preliminaries}
\label{section-2}

\begin{definition}
    \textbf{Dynamic Graph}. We consider a student solving a problem as an interaction between the student and the question and we represent the knowledge tracing dynamic graph as a sequence of interactions between students and questions.
    Let $\mathcal{G} = \{(s_1, q_1, t_1,r_1,k_1),$ $  (s_2, q_2, t_2,r_2,k_2),$ $ \cdots (s_I, q_I, t_I,r_I,k_I) \}$ be the set of interaction links, $0 \textless t_1 \leq  t_2 \leq \cdots t_I$, where $s_i \in \mathcal{S}$ denotes the student node and $q_i \in \mathcal{Q}$ denotes the question node of the $i$-th interaction at timestamp $t_i$. $I$ is the length of the interaction set.
   $\mathcal{S}$ is the set of all students, $\mathcal{Q}$ is the set of all questions.
    
  Each node $s \in \mathcal{S}, q \in \mathcal{Q}$ can be associated with node feature $\textbf{x}_{sn} \in R^{d_N},\textbf{x}_{qn} \in R^{d_N}$, and each interaction $(s,q,t,r,k)$ has link feature  $\textbf{x}_e \in R^{d_E}$. $d_N$ and $d_E$ denote the dimensions of the node feature and link feature.
  Each link has a label $r_{s,q}^t \in \left\{0,1\right\}$, which represents the correctness of the answer student $s$ give to question $q$, where 0 means wrong and 1 means right; $k_i \in \mathcal{K}$ is the knowledge concept associated with question $q_i$, which is obtained from initial question node feature.
 \end{definition}
 \begin{definition}
    \textbf{Problem Formalization}.  Given the student node $s$, question node $q$, timestamp $t$, and historical interactions before $t$, $ \{(s’,q’,t’,r',k')|t’ < t\} $, we will predict performance score of the student $s$ on the specific question $q$ at time $t$ by the student's representation $h_s^t$ and the question's representation $h_q^t$ within dynamic graph structure.
    
    We validate the effectiveness of the learned representations via the performance prediction task.
    Given a student solving question dynamic graph $\mathcal{G}_t$, it is aimed to design a model to learn time-aware representations $h_s^t \in R^d$ for the student node $s$ and $h_q^t \in R^d$ for question node $q$ with the nodes historical interaction sequences $S^t_s = \{(s,q_i,t_i,r_i,k_i)|t_i < t\}$ and $S^t_v = \{(s_j,q,t_j,r_j,k_j)|t_j < t\}$. Finally predict the performance $\hat{y}_t$ with representations above for node pairs $(s,q)$ at timestamp $t$ whose real performance is the interaction label $r$ of the two nodes at timestamp $t$.
    
\end{definition}

\section{Methodology}
\label{section-3}
\begin{figure*}[htbp]
	\includegraphics[width=1.0\linewidth]{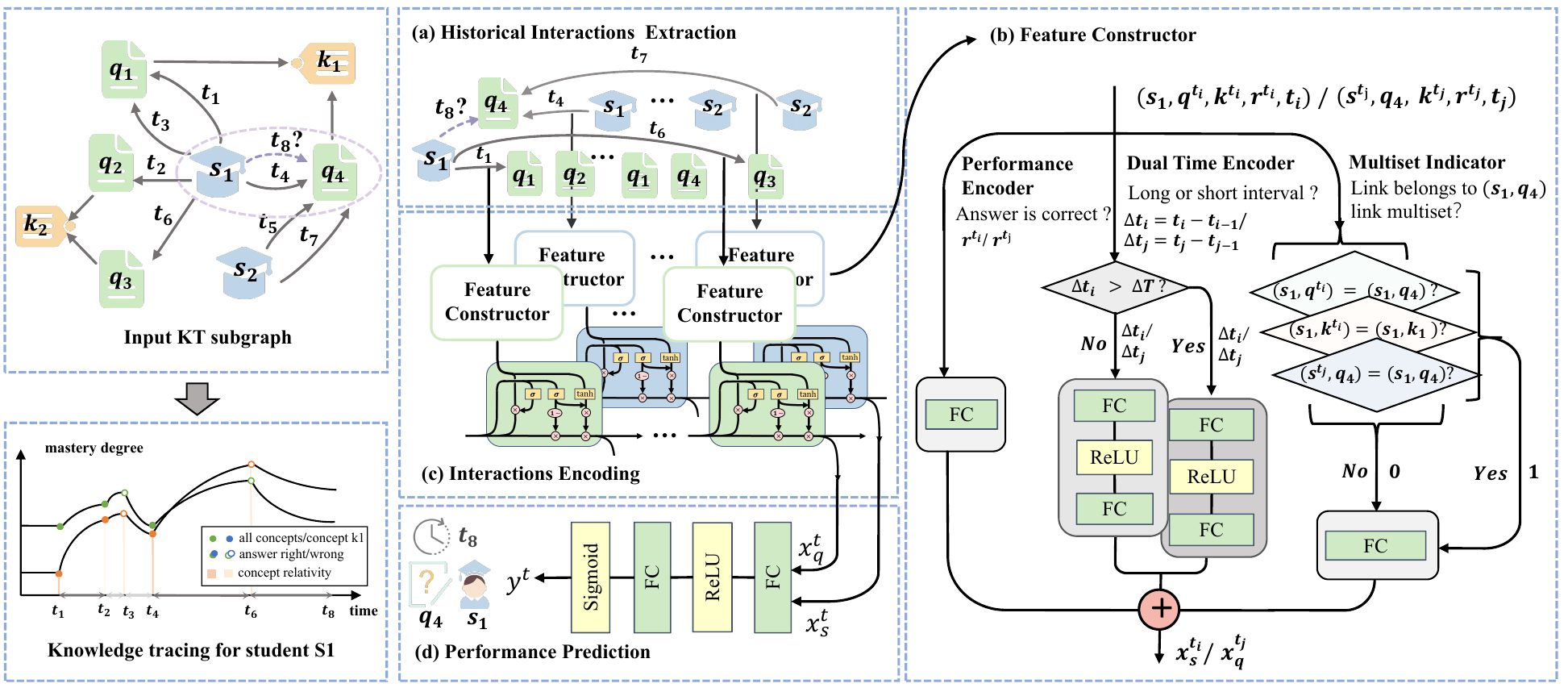}
    \caption{We develop a continuous dynamic graph learning model for KT. (a) We first convert input into a continuous dynamic graph, then extract the historical first-hop interaction of both the student node and the question node. (b) Next, embed the interaction neighbor sequence with a performance encoder, dual time encoder, and multiset indicator. (c) Then, we stack the two sequences with their respective features and feed them separately into two sequential models. (d) Finally, the outputs of the sequential models are used for the knowledge tracing task.}
\label{Fig:framework}
\vspace{-0.2cm}
\end{figure*}

This section presents the details of the Dynamic Knowledge Tracing Network(DyKT) The framework of DyKT is shown in figure \ref{Fig:framework}. Given an interaction $(s, q, t,r,k)$, we first extract historical first-hop interactions of student $s$ and question $q$ before timestamp $t$ and obtain two interaction sequences $S^t_s$ and $S^t_q$, which respectively represent the sequence of questions that the student $s$ has attempted and the records sequence of which students attempted the question q. Next, we encode performance, time intervals, and multiset information of student and question sequences. Then, we stack the two sequences with their respective features and feed them separately into two sequential models for capturing long-term temporal dependencies. Finally, the outputs of the sequential models are used in time-aware representations of $s$ and $q$ at timestamp $t$ (i.t., $h_s^t$ and $h^t_q$), which can be applied in knowledge tracing task.

\subsection{Subgraph Construction}
As we consider the relationship between students and questions as a graph structure, it is intuitive to understand that the first-order neighbor nodes of a student are all the questions that he/she has attempted, and the first-order neighbor nodes of a question are all the students who have attempted that question.

\textbf{Extract Historical Interactions. }In our model, 
we only need to extract a certain number of first-hop neighbors from historical records to establish sequences and learn from them. Mathematically, given an interaction $(s, q, t, r, k)$, for student node $s$ and question node $q$, we respectively obtain the sequences that involve the latest $N$ historical interactions of $s$ and $q$ before timestamp $t$, which are denoted by $S_s^t = \{(s,q',t',r',k')|t'<t\}$ and $S_q^t = \{(s',q,t',r',k)|t'<t\}$. To ensure consistent and efficient processing within a batch, we will use a patching technique to standardize the length of sequence that is less than $N$. This approach does not change the initial state of the student or the initial evaluation of the question but can accelerate the computation using a GPU tensor.

Now that we have obtained sequences $S_s^t$ and $S_q^t$ of length $N$, the next step is to identify the links, nodes, and structural features of interactions contained in these two interaction sequences. Based on these two sets of the most recent records before time t, we will compute the student's state representation at time t and assess the problem at time t.
    
\textbf{Performance Encoder. }In our KT model, the dynamic graph is associated with links whose labels have two types of visual representations. One is the link labels in the sequence $S^t_s$ based on the given performance of the student $s$  which are denoted by $\textbf{X}_{s,E}^t \in R^{N \times d_E}$, the other is link labels in the sequence $S^t_q$ based on the given score distribution of the question $q$ denoted by $\textbf{X}_{q,E}^t \in R^{N \times d_E}$. $\textbf{x}_{e}^{t_i}$ is one of the link feature in $\textbf{X}_{E}^t$ at time $t_i$, we generate the link feature by inherent label value $r^{t_i}$:
\begin{gather}
    \textbf{x}_{se}^{t_i} = \textbf{W}_{E}r_{s,q_i}^{t_i}+\textbf{b}_{E} ,\\
    \textbf{x}_{qe}^{t_j} = \textbf{W}_{E}r_{s_j,q}^{t_j}+\textbf{b}_{E},
\end{gather}
 where $r_{s,q_i}^{t_i}$ is the link label of student $s$ and $q_i$ at time $t_i$, $\textbf{W}_{E} \in R^{1 \times d_E}$, $\textbf{b}_{E} \in R^{d_E}$ are embedding matrix that need training.
 
\textbf{Dual Time Encoder. }The time interval between two consecutive problem-solving behaviors by a student exhibits a clear diversity. In our work, we distinguish the difference between long-term and short-term intervals. This indicates that students have two distinct learning patterns: one is continuous problem-solving behavior within a short period, and the other is resuming their learning after a certain interval. We believe that the learning and forgetting effects within a short-term interval differ from those in a long-term interval. Therefore, we propose an innovative dual-time model. 
In this dual-time module, we first calculate the time interval $\Delta t$ between a student's two consecutive problem-solving interactions $\Delta t_i = t_{i} - t_{i-1}$. Secondly, we classify a student's learning intervals between the previous and current submission records based on whether they were continuous within $\Delta T$, and then encode them according to the corresponding time interval type. If the interval between continuous learning is less than $\Delta T$, we model it by a short-term continuous learning model. For learning intervals that exceed $\Delta T$, we model and process them in the context of longer tutorials.
Specifically, a simple implementation approach is to use two different MLPs with different parameters for separate training of two types of periodic temporal patterns.
Encode timing of student $s$:
\begin{gather}
    \textbf{h}_{st}^{t_i}  = \begin{cases}
                \textbf{W}_{s}\Delta t_i + \textbf{b}_{s} & \text{if } \Delta t_i \leq \Delta T \\
                \textbf{W}_{l}\Delta t_i + \textbf{b}_{l} & \text{if } \Delta t_i > \Delta T
                    \end{cases} ,
\end{gather}
where
$\textbf{W}_{l} \in R^{1 \times d_T}$,
$\textbf{b}_{l} \in R^{d_T}$ are trainable matrix for long-term intervals embedding.
$\textbf{W}_{s} \in R^{1 \times d_T}$, 
$\textbf{b}_{s} \in R^{d_T}$ are trainable matrix for short-term intervals embedding.
As shown in the pie chart, there is a clear distinction in the distribution of intervals between less than 1 day and more than 1 day. We will now set the time interval, represented as $\Delta T$, to be 1 day.
The variable $\textbf{h}_{st}^{t_i} \in R^{d_T}$ represents the results of encoding different time intervals. Then, apply a non-linear layer, which is a Relu function in our case. $ \hat{\textbf{h}}_{st}^{t_i}=\textit{Relu}(\textbf{h}_{st}^{t_i})$, and further encode $\hat{\textbf{h}}_{st}^{t_i}$ uniformly to generate final time feature $\textbf{x}_{st}^{t_i}$, since there are still common patterns and features in all time interval information.

\begin{equation}
    \textbf{x}_{st}^{t_i} = \textbf{W}_T \hat{\textbf{h}}_{st}^{t_i} + \textbf{b}_T,
\end{equation}     
where $\textbf{x}_{st}^{t_i} \in R^{d_T}$, 
$\textbf{W}_{T} \in R^{d_T \times d_T}$,
$\textbf{b}_{T} \in R^{d_T}$ are the common parameters for all time intervals.
Similarly, we transfer and apply the same time encoding method to capture the underlying temporal information $\textbf{x}_{qt}^{t_j}$for dynamically assessing the difficulty levels and accuracy rates of questions.

\textbf{Multiset Indicator. }In our model, we encode the structural information of the graph by annotating whether the link in sequence belongs to the link multiset of the student $s$ and the question $q$ to be predicted. 
The practical meaning is to search and encode whether the question $q$ or its knowledge concept has been attempted by the student in sequence $S_s^t$ and whether student $s$ has previously attempted a particular question in sequence $S_q^t$. For all $(s,q_i,t_i,r_i,k_i) \in S_s^t$ and $(s_j,q,t_j,r_j,k_j) \in S_q^t$, perform the following calculations sequentially:

\begin{gather} 
    \textbf{h}_{snq}^{t_i}  = \begin{cases}
                1, & \text{if }  (s, q_i) = (s, q)\\
                0 , & \text{else}
                \end{cases},\\ 
    \textbf{h}_{snk}^{t_i}  = \begin{cases}
                1, & \text{if } (s, k_i) = (s, k)\\
                0 , & \text{else}
                    \end{cases},\\ 
    \textbf{h}_{qn}^{t_j}  = \begin{cases}
                1, & \text{if } (s_j,q) = (s,q)\\
                0 , & \text{else}
                    \end{cases},
\end{gather}
where $k_i, k_j \in \mathcal{K}$ is the knowledge concept associated with question $q_i, q_j$. We need to integrate the multiset features on questions $\textbf{h}_{snq}^{t_i}$ and knowledge concepts $\textbf{h}_{snk}^{t_i}$ for the student node. For problem nodes, the multiset feature on students $\textbf{h}_{qn}^{t_j}$ is directly used to calculate the node features later.
\begin{gather}                 
    \textbf{h}_{sn}^{t_i} = \textbf{h}_{snq}^{t_i} + \textbf{h}_{snk}^{t_i} ,\\
    \textbf{x}_{sn}^{t_i} = \textbf{W}_{m}\textbf{h}_{sn}^{t_i} + \textbf{b}_{m} ,\\
    \textbf{x}_{qn}^{t_j} =\textbf{W}_{m}\textbf{h}_{qn}^{t_j} + \textbf{b}_{m},
\end{gather}
where $\textbf{h}_{sn}^{t_i}$ is the multi-link information of student node $s$, $\textbf{x}_{sn}^{t_i} \in R^{d_N}$ is the representation of student's link multiset while $\textbf{x}_{qn}^{t_j} \in R^{d_N}$ is the representation of the question's link multiset.
$\textbf{W}_{m} \in R^{1 \times d_N}$,
$\textbf{b}_{m} \in R^{d_N}$ are trainable parameters.

\subsection{Student Knowledge Tracing}

The features of the student's sequence of question-solving events $X_s \in R^{N \times d_N}$ integrates the student's historical performance, their evolving learning or forgetting states over time as well as information about the relevance (position) to the solved questions or corresponding knowledge concepts. Additionally, we incorporate the inherent knowledge concept features of the questions. After obtaining the features of events sequence $X_s \in R^{N \times d_N}$, we aim to track the student's learning state using these features.

To achieve this, we utilize a continuous-time GRU model. 
GRU can selectively forget and update the new event information with the existing state, then effectively update and refine the student's knowledge state. Ultimately, it enables us to have a comprehensive understanding of the student's evolving knowledge state.

In this process, we input the event embedding $\textbf{x}_{s}^{t_i}$ into continuous-time GRU, which takes into account the previous knowledge state $\textbf{h}_{s}^{i-1}$ to compute the updated knowledge state $\textbf{h}_{s}^{i}$:

\begin{gather}
    \textbf{x}_{s}^{t_i} = \textit{Integ}_1(\textbf{x}_{se}^{t_i},\textbf{x}_{st}^{t_i}, \textbf{x}_{sn}^{t_i},\textit{Emb}(k^{t_i})) ,\\
    \textbf{o}_{s}^{i},\textbf{h}_{s}^{i} = \textit{GRUcell}_1(\textbf{x}_{s}^{t_i}, \textbf{h}_{s}^{i-1}) ,\\
    \textbf{x}_{s}^{t} = \textbf{W}_{out}\textbf{h}_{s}^{N} + \textbf{b}_{out},
\end{gather}
where $\textit{Emb}(.)$ output's dim is $d_E$, and $\textit{Integ}_1(.)$ is a Linear function that can integrate 
 $\textbf{x}_{se}^{t_i},\textbf{x}_{st}^{t_i}, \textbf{x}_{sn}^{t_i},\textit{Emb}(k^{t_i})$ into one embedding $\textbf{x}_{s}^{t_i} \in R^{d_E}$, 
$\textbf{W}_{out} \in R^{d_E \times d_E}$,
$\textbf{b}_{out} \in R^{d_E}$,
student's  knowledge state embedding is $\textbf{x}_{s}^{t} \in R^{d_E}$.

\subsection{Question Difficulty Assessment}
Similar to the process of tracking the student's learning state, we will also incorporate a dynamic assessment of question difficulty, rather than ID-embedding in existing methods to deal with unseen questions in the training data. We try to address the challenge of not having prior knowledge of the average accuracy or difficulty of the questions from the beginning. Besides, we seek to uncover the relationship between time and difficulty assessment. Input the question's sequence of events embedding $X_q \in R^{N \times d_N}$ into continuous-time GRU, which takes into account the previous difficulty assessment $\textbf{h}_{q}^{i-1}$ to compute the updated assessment $\textbf{h}_{q}^{i}$ and integrate the final difficulty assessment with the inherent knowledge concept features to encode the question.

\begin{gather}
    \textbf{x}_{q}^{t_j} = \textit{Integ}_2(\textbf{x}_{qe}^{t_j},\textbf{x}_{qt}^{t_j}, \textbf{x}_{qn}^{t_j}),\\
    \textbf{o}_{s}^{j},\textbf{h}_{s}^{j} = \textit{GRUcell}_2(\textbf{x}_{s}^{t_j}, \textbf{h}_{s}^{j-1}),\\
    \textbf{x}_{q}^{t} = \textbf{W}_{out}(\textbf{h}_{s}^{N}+\textit{Emb}(k^{t})) + \textbf{b}_{out},
\end{gather}
$\textit{Emb}(.)$ output's dim is $d_E$, and $\textit{Integ}_2(.)$ is a Linear function that can integrate 
 $\textbf{x}_{qe}^{t_j},\textbf{x}_{qt}^{t_j}, \textbf{x}_{qn}^{t_j}$ into one embedding $\textbf{x}_{q}^{t_j} \in R^{d_E}$, question's  assessment embedding is $\textbf{x}_{q}^{t} \in R^{d_E}$.

\subsection{Link Classification and Parameter Learning}
With the latest knowledge state embedding $\textbf{x}_{s}^{t}$ of student $s$ and latest assessment embedding $\textbf{x}_{q}^{t}$ of question $q$, we will predict student's performance on the question $q$ assigned to the student at $t$, which is the Dynamic Link Classification task in the dynamic graph. Let the probability that indicates the likelihood the student can answer the question $q$ correctly be written formally as: 
\begin{equation}
y^t = \textit{F}(\textbf{x}_{s}^{t}, {x}_{q}^{t}),
\end{equation}
where $\textit{F}(.)$ is the interaction function between the student's embedding and the question's embedding.
Here, we adopt a two-layer MLP as the interaction function, taking the concatenation of two embeddings as the input and obtaining 1 dim output. 
The final prediction probability of the student $S$ can correctly answer the question $q$ is $\hat{y}^t = \textit{Sigmoid}(y^t)$, which is calculated through the sigmoid function. 

We optimize all parameters in DyGKT by a standard cross-entropy loss between the predicted probability $\hat{y}_t$ and the true response $r_{s,q}^t$,

\begin{equation}
\mathcal{L} = -\sum_t^{t_I} (r_{s,q}^t log \hat{y}_t +(1-r_{s,q}^t)log(1-\hat{y}_t).
\end{equation}

We use Adam as the optimizing algorithm and add weight decay on factor matrices.

\section{Experiments}
\label{section-4}
\subsection{Datasets}
Our experiments utilize five publicly available datasets,  Table \ref{tab:datasets} shows the statistics of all datasets.  


\textbf{ASSISTment12}\footnote{\url{https://sites.google.com/site/assistmentsdata/datasets/2012-13-school-data-with-affect}} was collected from the ASSISTments online tutoring system. It contains data from skill-builder problem sets where students need to work on similar exercises to achieve mastery. The dataset includes data from the school year 2012-2013 and includes affect predictions. Records without knowledge concepts have been filtered out.

\textbf{ASSISTment17}\footnote{\url{https://sites.google.com/view/assistmentsdatamining/dataset}} was published in the 2017 ASSISTments Longitudinal Data Mining Competition. 

\textbf{Slepemapy.cz}\footnote{\url{https://www.fi.muni.cz/adaptivelearning/?a=data}} comes from an online adaptive system for practicing geography facts. It includes data on student interactions with the system, including exercises solved, mistakes made, and time spent on each exercise.

\textbf{Junyi}\footnote{\url{https://pslcdatashop.web.cmu.edu/DatasetInfo?datasetId=1198}} was gathered from Junyi Academy, an e-learning platform established in 2012. It contains data on student interactions with the platform, including exercises solved and time spent on each exercise.

\textbf{EdNet-KT1}\footnote{\url{https://github.com/riiid/ednet}} is the collection of all student-system interactions over two years by Santa, a multi-platform AI tutoring service with more than 780K users in Korea. It consists of students' exercise-solving logs. If an exercise has more than one knowledge concept, only the first knowledge concept is used for modeling.





\begin{figure}[t]
\centering
\includegraphics[width=0.90\linewidth]{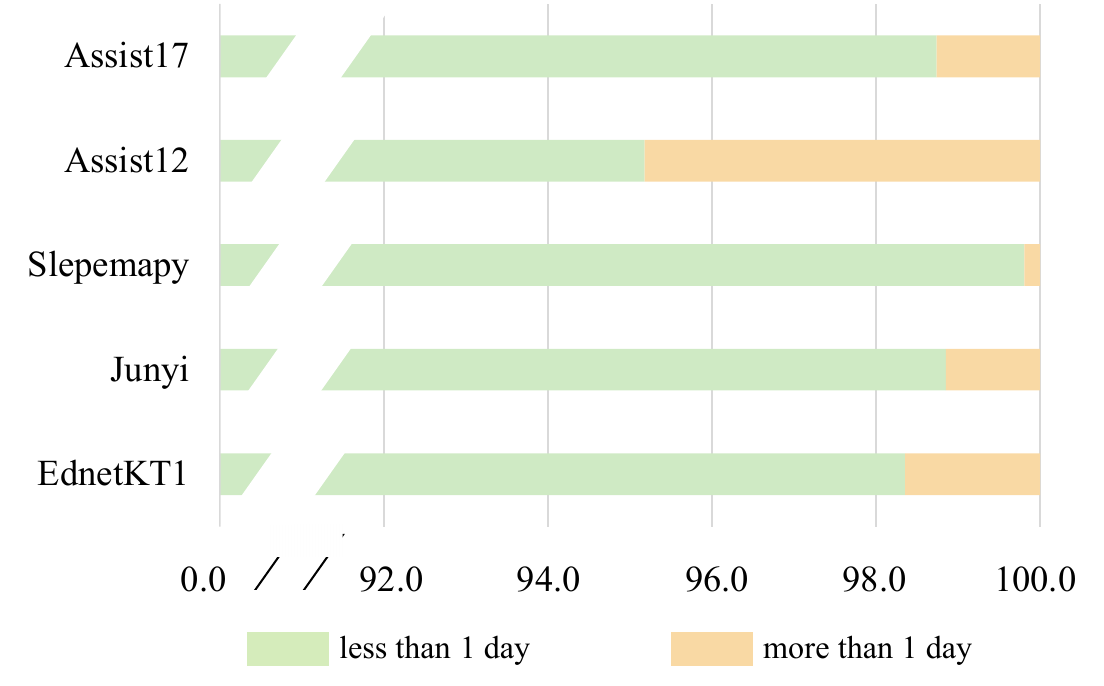}
\caption{Bar chart depicting the distribution of statistical time intervals for five datasets. It can be found that there are a large number of students answering questions continuously in the real data, and then starting another question after some time between breaks, which once again confirms the logic of our dual time encoder module.}
\label{Fig:count}
\vspace{-0.5cm}
\end{figure}

We remove records belonging to the students that have answered the questions less than 5 times, as well as the questions that have been answered less than 5 times to ensure that each sequence has sufficient data for cognitive modeling, following to Ma \cite{ma2022reconciling}.  Figure \ref{Fig:count} depicts the distribution of statistical time intervals for five datasets.  Detailed descriptions of the datasets are listed in Appendix \ref{sec:datasetdes}.

\begin{table}[h]
\caption{Statistics of all datasets.}
\label{tab:datasets}
\begin{tabular}{@{}lllll@{}}
\toprule
Dataset      & Student & Question & Concept & Interaction \\ \midrule
ASSISTment12 & 25.3k   & 50.9k      & 245   & 2, 621.3k   \\
ASSISTment17 & 1.7k    & 3.2k      & 102    & 942.8k      \\
Slepemapy.cz & 81.7k   & 2.9k   & 1, 458    & 9, 786.5k   \\
Junyi        & 175.4k  & 0.7k         & 40    & 25, 670.2k  \\
EdNet-KT1    & 685.4k  & 12.3k      & 141   & 95, 023.7k  \\ \bottomrule
\end{tabular}
\end{table}

\subsection{Evaluation}

\subsubsection{Baseline Models}

When comparing different approaches, we explore various methods in the Knowledge Tracing (KT) field and translate the sequence model into a continuous time dynamic graph framework to predict unlimited training data.

Baselines include classic models based on recurrent neural networks with the help of additional information or encoder, DKT\cite{piech2015deep}, IEKT\cite{long2021tracing}, LPKT\cite{shen2021learning}, DIMKT\cite{shen2022assessing}, CT-NCM\cite{ma2022reconciling}, and QIKT\cite{chen2023improving}. Also, there are attention-based models such as AKT\cite{pandey2019self} and simpleKT\cite{liu2023simplekt}. Apart from traditional KT models, We conduct tests under general dynamic graph frameworks, TGN\cite{DBLP:journals/corr/abs-2006-10637}, TGAT\cite{xu2020inductive}, DyGFormer\cite{yu2023towards}, and DyGKT to confirm the effectiveness of defining paradigms for training tasks under dynamic graphs and expressing dynamic features in the knowledge tracing task. Detailed descriptions of the baselines are listed in Appendix \ref{sec:baselinedes}.

\subsubsection{Evaluation Metrics}

 Regarding evaluation metrics, we employed Average Precision (AP) and the area under the receiver operating characteristics curve (AUC) to assess the performance of all methods in predicting binary-valued future student responses to exercises. AP measures the proportion of correctly predicted responses among all responses, while AUC reflects the ability of a model to distinguish between positive and negative responses. Additionally, we introduced an inductive sampling method to test the accuracy of students in answering questions that they have never encountered before. This sampling method involves hiding the nodes that will be used for inductive testing during the initial training phase, and then providing those test question nodes during the testing phase. The ratio of our inductive and transductive experiments for training, validation, and test sets is 0.8, 0.1, and 0.1.

\subsubsection{Model Configurations}
For existing KT model baselines designed based on fixed-length sequences, we propose a method that utilizes a dynamic graph. Firstly, we extract the current interaction between the student and the question, which represents the node pair to be predicted. Then, we retrieve the 50 historical neighbors of the student node following DKT\cite{piech2015deep}, which correspond to the N previous questions attempted by the student. These neighbors are then fed into the sequential model, and the output sequence's last value is obtained. This approach allows us to extend the model to an infinite-length sequence.


We conducted the experiments three times and took the average and the standard deviation of the results. However, for larger datasets and computationally intensive methods, only one run was performed, resulting in a standard deviation of 0. 


\subsubsection{Implementation Details}
We employed the Adam optimizer with a learning rate of 0.01 in all our experiments. The batch size was fixed at 2000 for training, validation, and testing. The model's hidden dimension is fixed to be 64. The dropout rate was set at 0.1 following TGN\cite{DBLP:journals/corr/abs-2006-10637}  for all experiments.  All experiments were implemented with PyTorch by Python and conducted with an Intel(R) Xeon(R) Gold 6130 CPU @ 2.10GHz having 16 physical cores and a GPU device NVIDIA Tesla T4 with a memory capacity of 12 GB. Our experiment codes are written based on DyGLib\footnote{\url{https://github.com/yule-BUAA/DyGLib}}, we will release the code when the paper is accepted.

\subsection{Performance Comparison}

\begin{table*}[ht]
\caption{AP and AUC-ROC for transductive and inductive prediction.}
\label{tab:t1}
\resizebox{\textwidth}{!}{
\centering
\begin{tabular}{@{}c|c|cccccccccc@{}}
\toprule
Setting            & Datasets  & \multicolumn{2}{c}{ASSITment12}   & \multicolumn{2}{c}{ASSITment17}   & \multicolumn{2}{c}{Slepemapy.cz}  & \multicolumn{2}{c}{junyi}         & \multicolumn{2}{c}{EdNet-KT1}     \\
\midrule
\multirow{15}{*}{Trans} & Metrics   & AP             & AUC             & AP             & AUC             & AP             & AUC             & AP             & AUC             & AP             & AUC             \\
\cline{2-12} 
                    & DKT       & 80.71 ± 0.73 & 68.00 ± 0.81 & 60.86 ± 0.62 & 70.98 ± 0.29 & 83.94 ± 0.26 & 64.62 ± 0.53 & 90.70 ± 0.09 & 69.01 ± 0.38 & 73.85 ± 0.51 & 62.48 ± 1.06 \\
                               & IEKT      & 83.83 ± 0.00 & 73.42 ± 0.01 & 63.93 ± 0.31 & 73.10 ± 0.26 & 87.00 ± 0.06 & 69.87 ± 0.13 & 92.76 ± 0.06 & 77.09 ± 0.12 &    76.41 ± 0.13 & 66.62 ± 0.12             \\
                               & LPKT      & 79.50 ± 0.87 & 67.64 ± 1.44 & 57.27 ± 0.07 & 68.25 ± 0.12 & 81.92 ± 5.46 & 62.32 ± 6.46 & 89.27 ± 0.18 & 65.76 ± 0.59 & 66.88 ± 0.12                &        55.48 ± 0.23         \\
                               & DIMKT     & 80.88 ± 0.16 & 69.14 ± 0.20 & 53.95 ± 0.14 & 64.45 ± 0.32 & 84.03 ± 0.01 & 68.09 ± 0.01 & 91.02 ± 0.01 & 72.51 ± 0.06 &        -        &        -        \\
                               & CT-NCM    & 80.33 ± 0.19 & 67.45 ± 0.28 & 59.74 ± 0.32 & 70.03 ± 0.22 & 84.20 ± 1.00 & 66.02 ± 0.33 & 90.91 ± 0.13 & 70.28 ± 0.57 & 75.00 ± 0.08 & 64.15 ± 0.12 \\
                               & QIKT      & 80.26 ± 0.03 & 69.11 ± 0.03 & 62.16 ± 0.35 & 71.80 ± 0.12 & 86.97 ± 0.07 & 69.79 ± 0.13 & 92.09 ± 0.05 & 75.38 ± 0.15 & 76.58 ± 0.11 & 66.64 ± 0.08 \\
                               & AKT       & 79.03 ± 2.57 & 65.57 ± 3.65 & 59.66 ± 0.10 & 68.29 ± 0.26 & 85.59 ± 0.25 & 67.39 ± 0.41 & 91.72 ± 0.20 & 73.10 ± 0.61 & 75.35 ± 0.15 & 64.68 ± 0.24 \\
                               & simpleKT  & 83.44 ± 0.81 & 72.78 ± 1.22 & 58.65 ± 1.06 & 66.67 ± 0.66 & 86.51 ± 0.06 & 68.62 ± 0.10 & 92.84 ± 0.07 & 76.98 ± 0.15 & 76.03 ± 0.09 & 66.60 ± 0.10 \\
                               & TGN       & \underline{86.25 ± 0.00}         & \underline{76.86 ± 0.00}         & 64.54 ± 0.38 & 72.83 ± 0.62 &   \underline{88.16 ± 0.08}       &    \underline{72.29  ± 0.11}   &        93.58 ± 0.17       &    78.91  ± 0.15        & \underline{81.94 ± 0.00}        & \underline{73.38 ± 0.00}        \\
                               & TGAT      & 85.90 ± 0.02 & 76.59 ± 0.04 & 63.24 ± 0.36 & 70.96 ± 0.41 & 85.24 ± 0.00         & 66.86 ± 0.00         &        93.67 ± 0.05         &           79.04 ± 0.18      & 77.23 ± 0.63 & 68.23 ± 0.53 \\
                               & DyGFormer & 86.03 ± 0.19 & 76.67 ± 0.29 & \underline{68.17 ± 0.14} & \underline{77.48 ± 0.12} & 87.42 ± 0.16 & 70.49 ± 0.24 & \underline{93.77 ± 0.03}                &    \underline{79.66 ± 0.02}             & 81.26 ± 0.08 & 72.57 ± 0.10 \\
                               & DyGKT     & \textbf{87.07 ± 0.30} & \textbf{78.04 ± 0.42} & \textbf{71.81 ± 0.52} & \textbf{80.21 ± 0.37} & \textbf{88.39 ± 0.04} & \textbf{72.49 ± 0.09} & \textbf{94.01 ± 0.01} & \textbf{80.35 ± 0.02} & \textbf{83.10 ± 0.03} & \textbf{74.33 ± 0.07} \\
\cmidrule(l){1-12} 
\multirow{15}{*}{Ind}    & Metrics   & AP             & AUC             & AP             & AUC             & AP             & AUC             & AP             & AUC     & AP             & AUC        \\
\cline{2-12}  
                    & DKT       & 80.10 ± 0.62 & 66.89 ± 0.68 & 60.56 ± 0.62 & 70.85 ± 0.30 & 83.87 ± 0.25 & 64.57 ± 0.53 & 90.28 ± 0.12 & 67.22 ± 0.39 & 72.97 ± 0.55 & 61.63 ± 1.15 \\
& IEKT      & 83.52 ± 0.01 & 72.93 ± 0.01 & 63.76 ± 0.33 & 73.04 ± 0.28 & 86.94 ± 0.06 & 69.82 ± 0.13 & 92.78 ± 0.07 & 76.82 ± 0.14 & 75.79 ± 0.13 & 66.11 ± 0.10 \\
& LPKT      & 79.82 ± 0.87 & 67.99 ± 1.44 & 57.28 ± 0.05 & 68.34 ± 0.12 & 81.86 ± 5.46 & 62.28 ± 6.43 & 89.08 ± 0.19 & 64.93 ± 0.56 & 66.83 ± 0.18 & 55.67 ± 0.25 \\
& DIMKT     & 80.35 ± 0.14 & 68.34 ± 0.19 & 53.70 ± 0.15 & 64.30 ± 0.33 & 83.98 ± 0.01 & 66.52 ± 0.00 & 90.80 ± 0.01 & 71.84 ± 0.06 &        -        &        -        \\
& CT-NCM    & 79.68 ± 0.19 & 66.27 ± 0.29 & 59.35 ± 0.31 & 69.84 ± 0.22 & 84.13 ± 1.01 & 65.97 ± 0.33 & 90.52 ± 0.20 & 68.73 ± 0.80 & 74.07 ± 0.09 & 63.24 ± 0.13 \\
& QIKT      & 80.50 ± 0.03 & 69.39 ± 0.03 & 61.95 ± 0.35 & 71.72 ± 0.12 & 86.92 ± 0.07 & 69.74 ± 0.13 & 92.08 ± 0.06 & 75.14 ± 0.16 & 75.98 ± 0.11 & 66.17 ± 0.08 \\
& AKT       & 78.37 ± 2.67 & 64.35 ± 3.87 & 59.39 ± 0.11 & 68.12 ± 0.27 & 85.53 ± 0.25 & 67.33 ± 0.41 & 91.46 ± 0.22 & 71.94 ± 0.71 & 74.48 ± 0.18 & 63.87 ± 0.29 \\
& simpleKT  & 83.25 ± 0.72 & 72.45 ± 1.06 & 58.43 ± 1.03 & 66.52 ± 0.64 & 86.45 ± 0.06 & 68.55 ± 0.10 & 92.91 ± 0.08 & 76.86 ± 0.17 & 75.39 ± 0.08 & 66.10 ± 0.09 \\
& TGN       & \underline{86.36 ± 0.00}        & \underline{77.06  ± 0.00}        & 64.57 ± 0.40 & 72.89 ± 0.63 &   \underline{88.12 ± 0.07}       &    \underline{72.25  ± 0.12}   &        93.84 ± 0.12       &    79.41  ± 0.18            & \underline{81.85 ± 0.00}         & \underline{73.41  ± 0.00}        \\
& TGAT      & 86.03 ± 0.01 & 76.82 ± 0.04 & 63.18 ± 0.37 & 70.95 ± 0.43 & 85.17 ± 0.00        & 66.80 ± 0.00         &       93.88 ± 0.04        & 79.36 ± 0.17     & 77.07 ± 0.64 & 68.24 ± 0.55 \\
& DyGFormer & 86.23 ± 0.17 & 76.98 ± 0.25 & \underline{68.22 ± 0.13} & \underline{77.59 ± 0.12} & 87.36 ± 0.16 & 70.44 ± 0.23 & \underline{93.97 ± 0.03} & \underline{79.95 ± 0.04} & 81.16 ± 0.15 & 72.57 ± 0.10 \\
& DyGKT     & \textbf{87.19 ± 0.27} & \textbf{78.23 ± 0.38} & \textbf{71.86 ± 0.52} & \textbf{80.32 ± 0.38} & \textbf{88.34 ± 0.05} & \textbf{72.45 ± 0.10} & \textbf{94.19 ± 0.01} & \textbf{80.62 ± 0.01} & \textbf{83.00 ± 0.03} & \textbf{74.34 ± 0.08} \\

\cmidrule(l){1-12} 
\end{tabular}
}
\vspace{-0.3cm}
\end{table*}

\textbf{Result}. Table \ref{tab:t1} presents the performance of our DyGKT model and other comparison approaches under the dynamic graph framework in predicting future student performance across five datasets. The reported results are averages obtained from three rounds of experiments. Optimal results are highlighted in bold, while sub-optimal results are marked in italics and underlined.
Please note that DIMKT needs to use the accurate rate of the concept of one question and it cannot process a dataset whose question has multiple concepts, so its results under EdnetKT1 are not presented. TGN, TGAT, and DyGFormer consume a considerable amount of time, and in some datasets, they need nearly a day for each round of computation on some datasets, so some of the results present only one round of experiments. The cost of time is in Appendix \ref{sec:appendix_exp}.

The visualization of a student's knowledge tracing of one specific question over the learning process is depicted in figure \ref{Fig:case2}, and more cases are in Appendix \ref{sec:casestudy}.

\begin{figure*}[!htbp]
\centering
\includegraphics[width=1.0\linewidth]{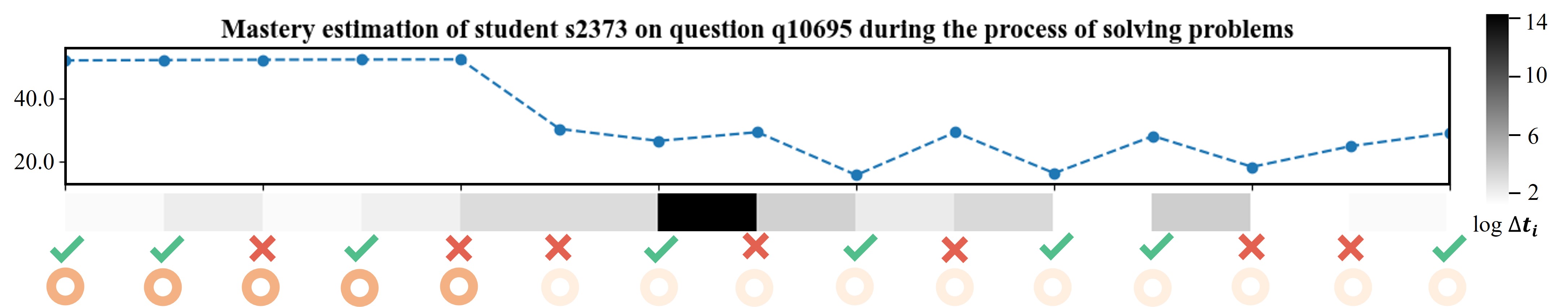}
\caption{Visualization of a student's knowledge mastery degree of one question over 15 steps in Assist17; the boxes with different colors in the time interval line represent different time interval ranges between the current and next steps; the darker-colored circles indicate that the attempted question belongs to the predicted link multiset.}
\label{Fig:case2}
\vspace{-0.3cm}
\end{figure*}

The DyGKT model has achieved the highest level of reliability among all models. This signifies that the classic dynamic graph models are capable of effectively handling the KT task within the dynamic graph and the dynamic graph structure can be extended to other sequence models. Other observations based on the table and results from previous work are as follows:

(1) There are differences compared to the results reported in the original methods due to the expansion of the test data. The original sequential approach only considers the initial 50 or 200 records of students for training and testing, and the model can track the initial state changes of students reasonably well. However, this also means that many models cannot adapt and model all learning states effectively. Particularly, as students progress and attempt a larger number of questions, the changes in their learning states become less easily captured and accurately assessed. Previous models struggle to fit this situation where time and question volume continue to increase indefinitely. For example, this change was reflected in the CTNCM paper\cite{ma2022reconciling}, where the CTNCM model performed best when the sequence length was 100 or 200, but accuracy decreased when increased to 500. Models based on DKT and AKT only exhibited slight improvements in accuracy within a range of 0.01.

(2) The inductive experiment also shows that if a student attempts a question they have never encountered before, the model's performance remains similar to the transductive effect. This indicates that previous KT models rely more on the concept features that are initially associated with the questions. When a new question appears, even if it contains knowledge concepts that have been encountered by the student in previous training, the model can still provide relatively accurate predictions.

(3) In the general framework of dynamic graphs, it has been found that previous models can still be effective when transferred to link classification tasks and exhibit outstanding performance. However, due to the increased computational processes and memory in dynamic graph models, the computation speed is significantly slowed down. This limits their ability to better adapt to large-scale student question prediction tasks.

\subsection{Ablation Study}
\label{sec:ablation}
We conducted an ablation study to validate the effectiveness of certain designs in DyGKT. The study involves examining the use of Multiset Indicator (MI), Dual Time Encoder (dtE), concept embedding (cE) in Question Difficulty Assessment, and time Encoding (tE). Besides, We encode questions by ID-embedding with either question ID or concept ID embedding but not Question Difficulty Assessment (qidE/cidE). We removed each module separately and referred to the remaining parts as w/o MI, w/o dtE, w/o tE, qidE, and cidE. We evaluated the transductive and inductive performance of different variants on the five datasets mentioned above and presented the results in figure \ref{Fig:ablation_trans}. The remaining experimental results are shown in the Appendix \ref{sec:appendix_exp}.
\begin{figure}[t]
\centering
\includegraphics[width=0.95\linewidth]{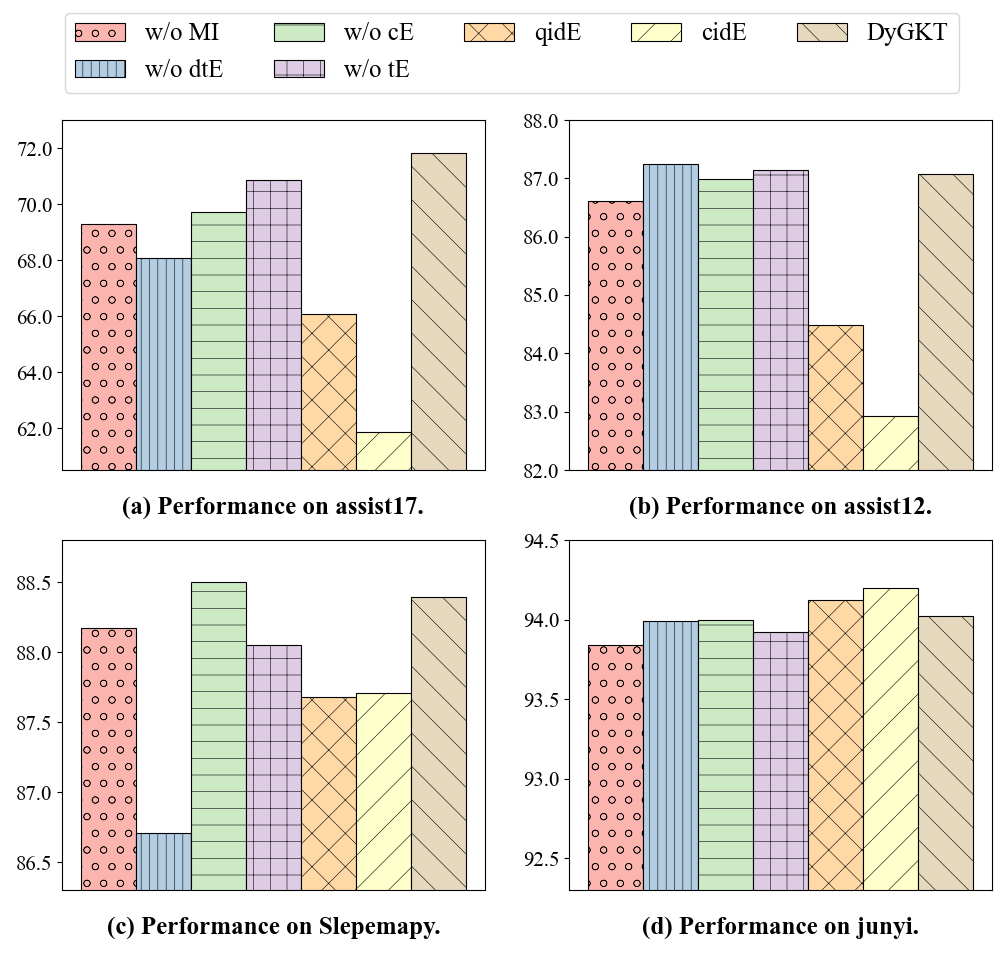}
\caption{AP for transductive ablation study of DyGKT.}
\label{Fig:ablation_trans}
\vspace{-0.5cm}
\end{figure}

Despite significant variations in the time distribution and the structure differences among different datasets, DyGKT exhibited stable and reliable performance across all datasets.
We found that the Question Difficulty Assessment had an unexpected effect when comparing qidE, kidE, and DyGKT. Dynamically assessing the questions not only helps predict performance but also assesses students' ability to solve questions and trace learning status.
The multiset indicator, shown by w/o MI, had a significant impact on performance. It effectively captured the structural encoding between student-question node pairs in the subgraph. It directly represented the student's mastery of the question they were attempting to answer and its corresponding concept via a multi-link dynamic graph.
Dual Time Encoder benefits model performance by generating short-term or long-term time interval features. According to the comparison with only a single-pattern time encoder (w/o dtE) and the model without time encoding (w/o tE), time encoding without diverse pattern distinguishing may even harm model performance on assist17 and Slepemapy.cz.
W/o cE showed a slight decrease, indicating that in DyGKT, concepts have a relatively stable impact on question representation but there is not a strong reliance on concepts to represent questions.


\subsection{Parameter Sensitivity}

\begin{figure}[H]
\centering
\vspace{-0.1cm}
\includegraphics[width=1\linewidth]{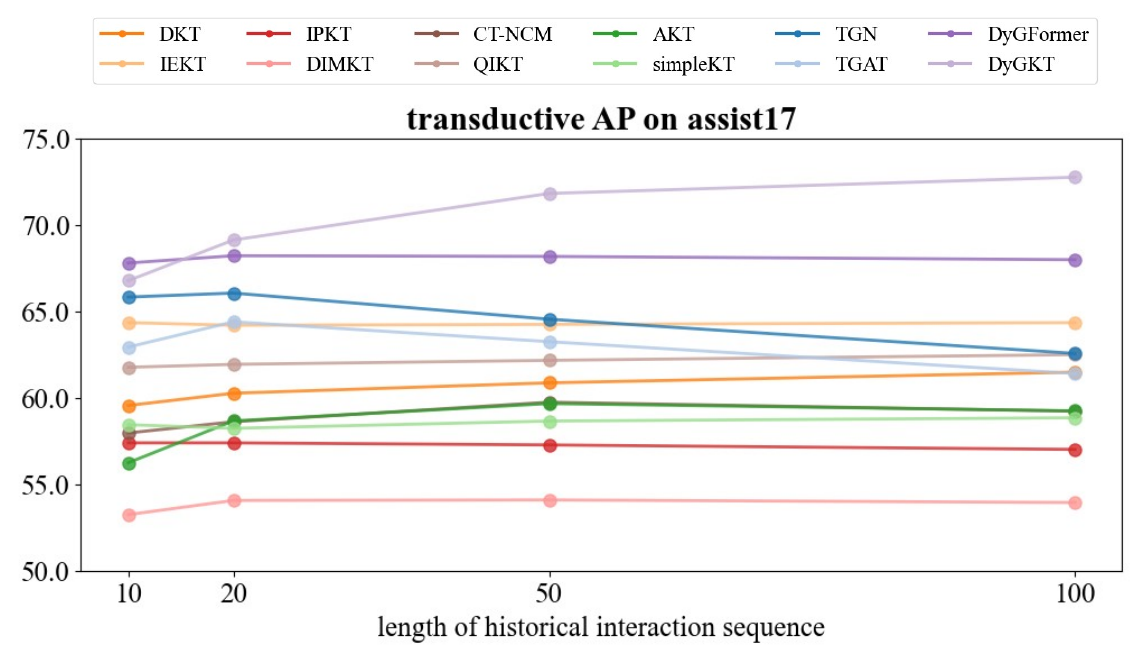}
\caption{Sequence length sensitivity in transductive setting.}
\label{Fig:sequence length}
\vspace{-0.2cm}
\end{figure}

\textbf{Length of the Sequence}. The length of the historical interaction sequence $S_s^t$ and $S_q^t$ is an important factor as it contains both valuable information from neighbors and noise and also affects the number of neighbor-induced subgraphs. In figure \ref{Fig:sequence length}, we test the sensitivity of the Sequence length parameter on the assist17 dataset for transductive setting, the results of the inductive sampling strategy are shown in Appendix \ref{sec:appendix_exp}.
The performance of the original dynamic graph model tends to deteriorate as the neighbor sequences become longer. This phenomenon may be attributed to the use of trigonometric functions for position time encoding, which fails to flexibly capture temporal features in the learning process. 
For classical KT baselines, there is a general trend of improving accuracy to some extent with the increase in sequence length, because longer sequences provide more information to track the students' learning states. Recurrent neural network (RNN) models commonly outperform attention-based models.

\subsection{Computation Cost Comparison} 

\begin{table*}[t]
\caption{Details and computation cost for Assist17 of all models.}
\label{tab:model cost}
\resizebox{\textwidth}{!}{
\centering

\begin{tabular}{@{}ccccccc@{}}
\toprule
KT Model  & Learning Model           & Time/Forgetting Encoder & Model Parameter(MB) & GPU Usage(MB) & Training Time(s) & Range in Assist17 \\ \midrule
DKT       & RNN/LSTM                 & -                       & 6.483               & 47.952        & 18.873           & 7                 \\
IEKT      & GRU                      & -                       & 12.452               & 52.169        & 124.204          & 4                 \\
LPKT      & GRU                      & MLP                     & 0.331               & 62.535        & 213.820          & 11                \\
DIMKT     & GRU                      & -                       & 3.478               & 20.676        & 17.077           & 12                \\
CT-NCM    & LSTM                     & Hawkes LSTM                    & 12.978               & 87.781         & 27.227            & 8                 \\
QIKT      & LSTM                     & -                       & 5.057                & 66.920         & 42.622            & 6                 \\
AKT       & Attention                & -                       & 10.080              & 165.781       & 62.864           & 9                 \\
simpleKT  & Attention                & -                      & 3.634               &  137.251      & 61.267          & 10                \\
TGAT      & Dynamic graph/Attention  & Fourier Time Encoding                & 0.410               & 13397.957         & 946.214          & 3                 \\
DyGFormer & Dynamic graph/Attention  & FourierTime Encoding                & 3.811               & 475.936        & 443.538          & 5                 \\
TGN       & Dynamic graph/Memory/GRU & Fourier Time Encoding                & 0.610               & 13412.284       &  2042.245            & 2                 \\
DyGKT     & Dynamic graph/GRU        & Dual MLP               & 0.292               & 55.307        & 47.400          & 1                 \\ \bottomrule
\end{tabular}
}
\end{table*}

We choose the dynamic graph structure rather than a sequential structure to model the learning process. Firstly, we sort all the question records in the dataset in ascending order of time. Each batch consists of a batch size of answering records, and each record contains a pair of nodes representing a student and a question. But present approaches consider a batch as a batch size of students, and perform predictions on their question-answering records as a whole, treating the sequence as static for training and prediction.

Setting node feature dimension/hidden dimension as $d$, neighbor sample length as $L$, node memory dimension as $M(M<d)$, number of edges as $E$, number of nodes as $N$, and number of student nodes as $S(S<N<E)$.

Our time complexity is $O(2*EL)$, and spatial complexity is $O(Ed+NM)$. Because we need to predict each pair by the previous $L$ records. Factor 2 is due to encoding calculations performed on both the neighbor sequences of the student node and the question node, as we don't present questions by embedding techniques. 

In the traditional static KT models, $(L-1)$ predictions are made in the sequence models once the student's $L$ interactions are put in. So the time complexity of the traditional static KT models is $O(SL)$, and the space complexity is $O(Ed+Nd)$. 

But we will also compare the traditional KT models when they are implemented within the dynamic graph. We expand the original KT methods to predict the pair of the student and the question based on the current student's historical answering sequence, and the sequence length maintains L. The time complexity for this calculation is $O(ELd)$, and the space complexity is $O(Ed+Nd)$.

The exact computation time of the model varies depending on the specific encoding method employed. For example, the AKT model utilizes attention mechanisms for computation, while the DKT model employs LSTM. The size of the model can measure the computational intensity. We have compiled a comparison of the methods used by all models in our paper, along with their respective time and space complexities details in Table \ref{tab:model cost}.

\section{Conclusion}
\label{section-6}

In this paper, we present DyGKT, a novel approach to knowledge-tracing modeling that utilizes continuous time dynamic graph networks. Our approach takes into account the dynamic nature of scale, time, and structure, allowing us to model the underlying graph that describes students’ learning processes. We frame the problem of knowledge tracing as a dynamic node classification challenge on a graph that is continuously evolving. This means that we can evaluate the performance of our model on graphs of infinitely growing length, rather than just fixed-length sequences. To capture the various temporal patterns within diverse interaction time intervals, we propose a dual time encoder. Additionally, we introduce the multiset indicator to leverage the graph’s structure, which reflects the inherent structural features in connections among students, questions, and concepts. Finally, to validate the effectiveness of our approach, we conducted a comprehensive statistical analysis of the datasets, as well as comparative experiments on five real-world datasets and ablation studies. 
We empirically validate our approach, and the results not only demonstrate the superior performance of DyGKT but also highlight the capability to effectively model Knowledge Tracing within the dynamic graph.



\section{Acknowledgments}
This work was supported by the National Natural Science Foundation of China (62272023, 51991395, 51991391, U1811463), the S\&T Program of Hebei(225A0802D), the Science and Technology Development Fund (FDCT), Macau SAR (file no. 0123/2023/RIA2, 001/2024/SKL), and the Start-up Research Grant of University of Macau (File no. SRG2021-00017-IOTSC).

\newpage

\bibliographystyle{ACM-Reference-Format}
\bibliography{reference}

\clearpage
\appendix
\section{Appendix} 
\label{section-appendix}
In the appendix, details of the experiments are introduced.
\subsection{Datasets Analysis} 
\label{sec:datasetdes}
ASSISTment12 contains student exercise data from the 2012-2013 school year, sourced from the ASSISTments online tutoring service system. ASSISTment17 was featured in the 2017 ASSISTments Longitudinal Data Mining Competition. Slepemapy.cz is derived from the online adaptive system slepemapy.cz, designed for geography practice. Junyi is collected from Junyi Academy, an E-learning platform established in 2012. Finally, EdNet-KT1 is an artificial intelligence tutoring system aiding students in TOEIC preparation.

\begin{figure*}[b]
\centering
    \subfigure[maximum repeating times]
    {
    \begin{minipage}[b]{0.3\linewidth}
        \includegraphics[width=1.0\linewidth]{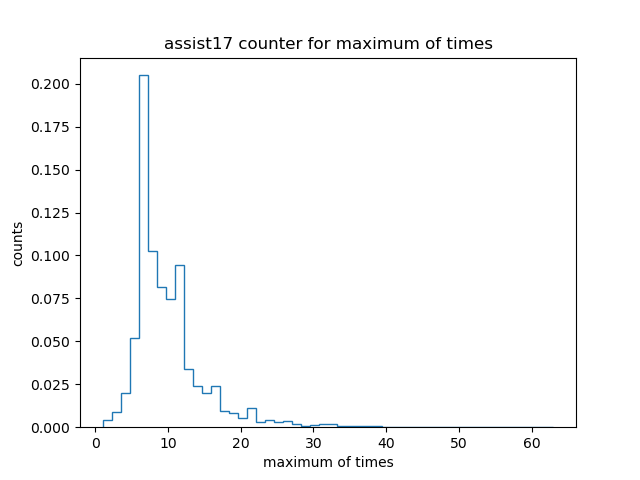}
    \end{minipage}
    }
    \subfigure[average repeating times]
    {
    \begin{minipage}[b]{0.3\linewidth}
        \includegraphics[width=1.0\linewidth]{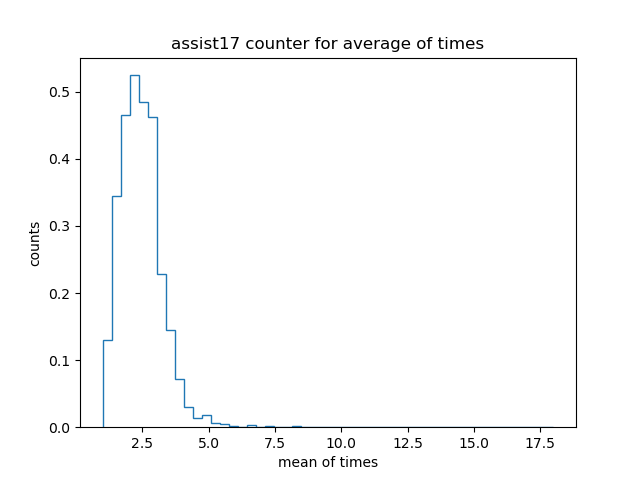}
    \end{minipage}
    }
    \subfigure[repeating times before success]
    {
    \begin{minipage}[b]{0.3\linewidth}
        \includegraphics[width=1.0\linewidth]{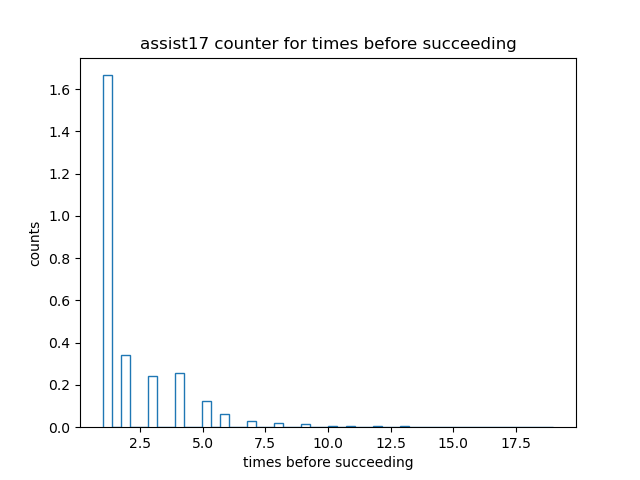}
    \end{minipage}
    }

\caption{Distribution of repeating times in assist17(Density).}
\label{Fig:assist17 repeating times}
\end{figure*}

We further explore the features of five real-world datasets. The following three columns represent 10,000 sampled pairs of (student, question) nodes. They include statistics on the maximum and average number of times students repeat questions and the average number of attempts students make before solving a question correctly. Take assist17 as an example, we observe that students exhibit a significant amount of repetition in solving questions in each dataset, as is shown in figure \ref{Fig:assist17 repeating times}. Additionally, the highly similar distributions across the three categories indicate that students often go through numerous failures before correctly answering a question. This underscores the significance of defining and identifying the behavior of repeating questions for predicting performance.



\subsection{Descriptions of Baselines}
\label{sec:baselinedes}
\textbf{DKT} model is the first knowledge tracing model that utilizes a recurrent neural network, specifically LSTM, to assess the knowledge state of students.

\textbf{AKT} model employes an attention network to capture the relevance between the knowledge components (KCs) and the historical interactions of students.

\textbf{LPKT} model directly models students' learning process to monitor their knowledge state.

\textbf{DIMKT} model explicitly incorporates the difficulty level into the representation of questions and establish the relationship between a student's knowledge state and the difficulty level of the questions during the practice process.

\textbf{IEKT} model estimates a student's cognition of a question before predicting their response. It also assesses the student's sensitivity to knowledge acquisition on the questions before updating their knowledge state.

\textbf{QIKT} is an interpretable knowledge-tracing model that focuses on individual questions. It estimates the variations in a student's knowledge state at a fine-grained level, using question-sensitive cognitive representations learned from a question-centric knowledge acquisition module and a question-centric problem-solving module.

\textbf{simpleKT} is a strong yet simple method for knowledge tracing. It models question-specific variations based on the Rasch model and utilizes an ordinary dot-product attention function to extract time-aware information embedded in a student's learning interactions.

\textbf{CT-NCM} model adapts the neural Hawkes process and proposes a learning function to model the change of a student's knowledge jointly states at each interaction. It naturally integrates dynamic and continuous forgetting behaviors into the modeling of the student's learning process with continuous time.

\textbf{TGN} model maintains an evolving memory for each node and updates this memory when the node is observed in an interaction. This is achieved through the message function, message aggregator, and memory updater. Additionally, an embedding module is utilized to generate temporal representations of nodes.

\textbf{TGAT} model computes the node representation by aggregating features from the temporal-topological neighbors of each node, using a self-attention mechanism. It also includes a time encoding function to capture temporal patterns.

\textbf{DyGFormer} model represents nodes with additional structural information. It extracts the first hop neighbor sequence of the end nodes and computes the neighbor co-occurrence encoding as the first-order relative position label.

The comparison of details of baselines and DyGKT is shown in Table \ref{tab:model cost}.

\subsection{Supplementary to Experiment Results} 
\label{sec:appendix_exp}



\begin{figure}[H]
\vspace{-0.2cm}
\centering
\includegraphics[width=1\linewidth]{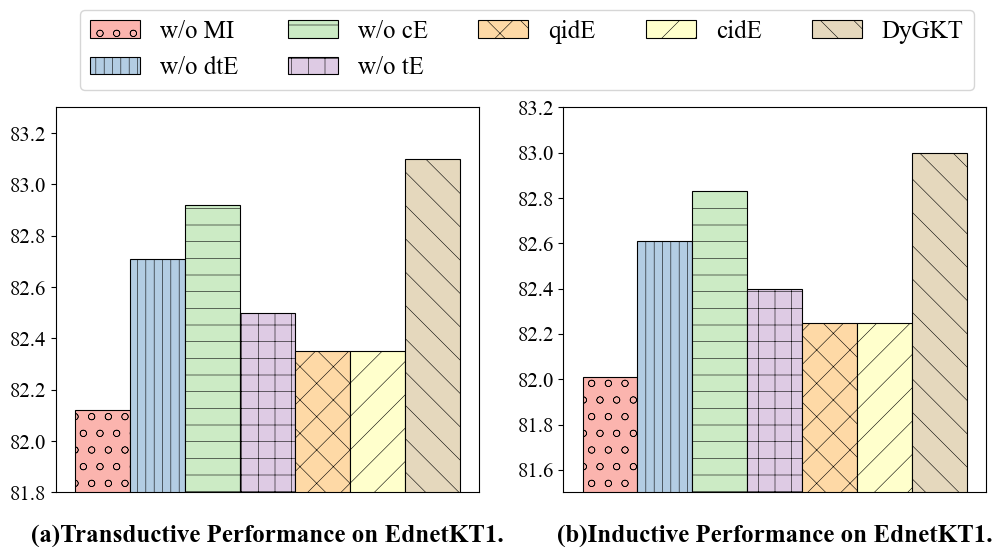}
\caption{AP in transductive and inductive setting for ablation study of the DyGKT in EdnetKT1}
\label{Fig:EdnetKT1 result}
\end{figure}

The ablation results of EdnetKT1 are revealed in figure \ref{Fig:EdnetKT1 result}. DyGKT performs the best on the EdNetKT1 dataset, while w/o MI, which performs well on the first four datasets, exhibits the poorest performance. Overall, when considering the performance across all five datasets, DyGKT emerges as the most stable and reliable model.

\begin{figure}[H]
\centering
\includegraphics[width=1\linewidth]{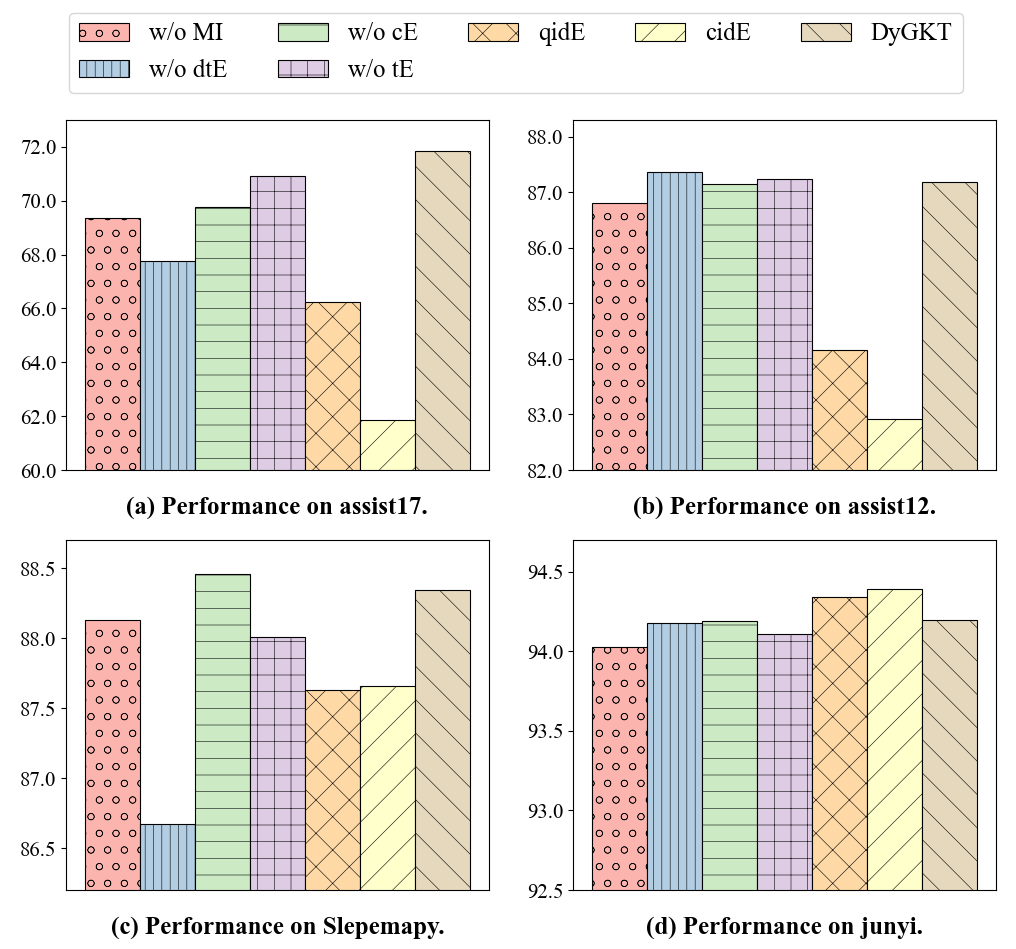}
\caption{AP for inductive ablation study of DyGKT.}
\label{Fig:ablation_in}
\end{figure}

Additionally, as is shown in figure \ref{Fig:ablation_in} comparing inductive experiments with transductive experiments reveals a consistent performance with a slight decrease of around 0.002. This indicates that the model adapts well to inductive scenarios as well.

\begin{figure}[h]
\centering
\includegraphics[width=1\linewidth]{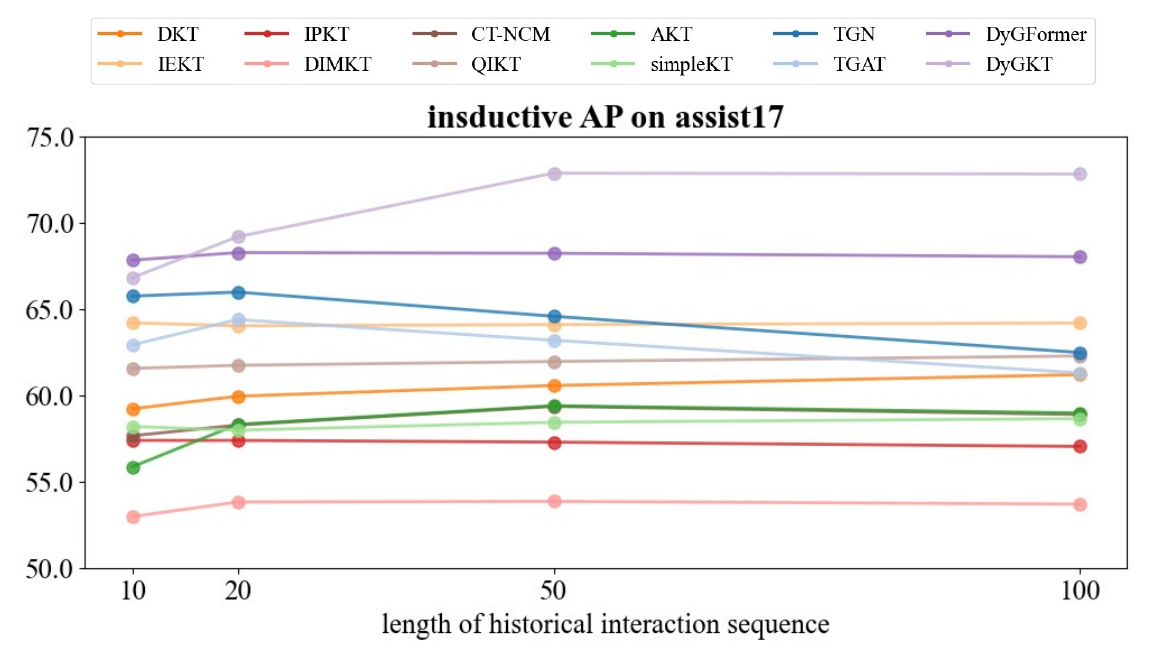}
\caption{Sequence length sensitivity in inductive setting.}
\vspace{-0.4cm}
\label{Fig:sequence length in}
\end{figure}

We examine the impact of the Sequence Length parameter on the assist17 dataset in an inductive setting. 
Similar results are observed under the transductive setting. Figure \ref{Fig:sequence length in} displays the results of the inductive sampling strategy. The original dynamic graph model tends to performance degradation as the neighbor sequences grow longer. This decline may be attributed to the use of position functions for time encoding, which inadequately captures temporal features during learning. In classical Knowledge Tracing (KT) tasks, longer sequences generally enhance accuracy by providing more information to track students' learning states. In recurrent neural network (RNN) tasks, longer sequences typically outperform attention-based tasks.

\subsection{case study}
\label{sec:casestudy}
\begin{figure}[!htbp]
\vspace{-0.2cm}
\centering
    \subfigure[assist17 $s_{819}$ learning process]
    {
    \begin{minipage}[b]{1\linewidth}
        \includegraphics[width=1.0\linewidth]{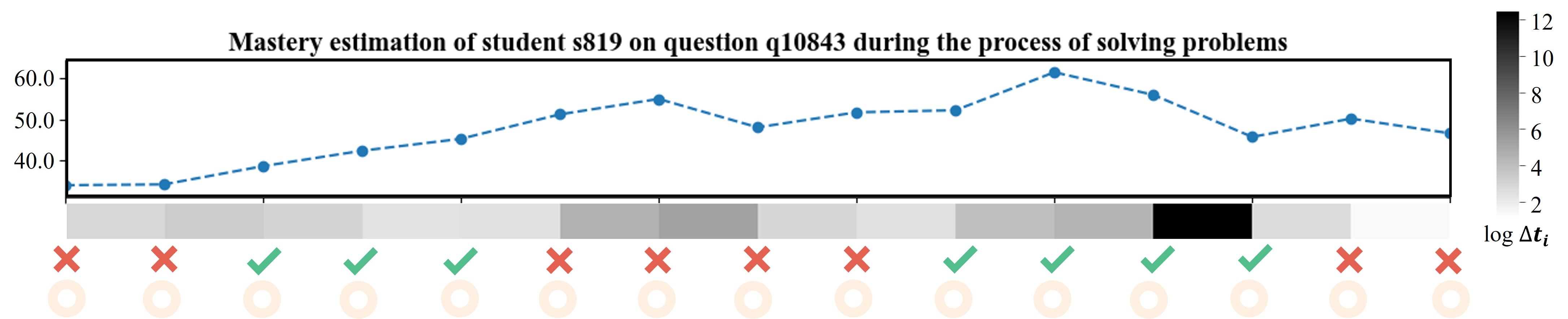}
    \end{minipage}
    }
    \subfigure[assist17 $s_{7776}$ learning process]
    {
    \begin{minipage}[b]{1\linewidth}
        \includegraphics[width=1.0\linewidth]{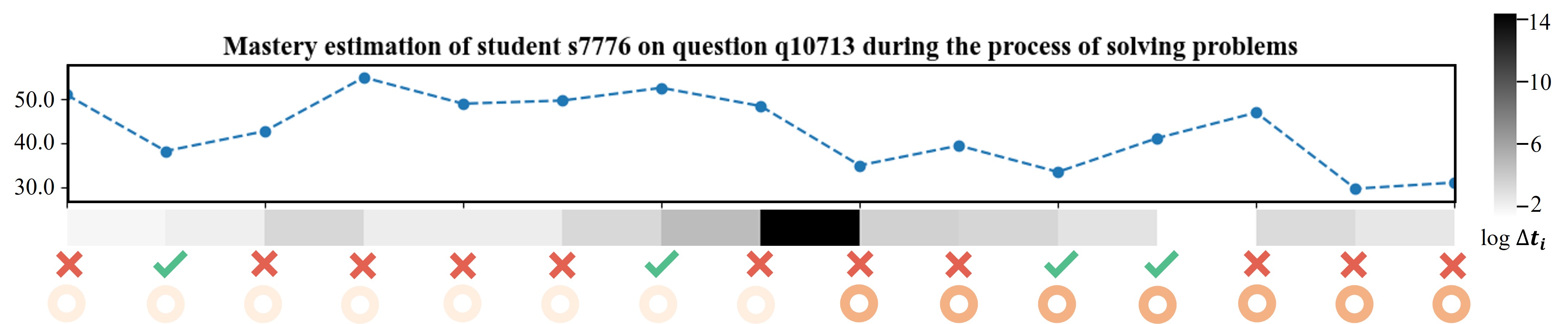}
    \end{minipage}
    }
    \subfigure[assist12 $s_{62118}$ learning process]
    {
    \begin{minipage}[b]{1\linewidth}
        \includegraphics[width=1.0\linewidth]{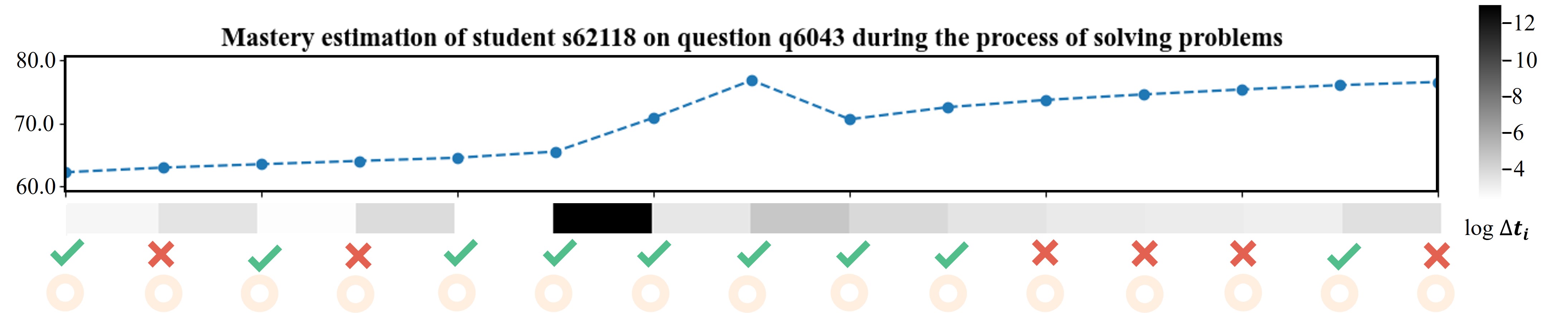}
    \end{minipage}
    }

\caption{Visualization of more students' knowledge mastery degree of questions over 15 steps.}
\label{Fig:case study ablation}
\end{figure}

We adopt a visualization approach to assess a student's mastery level of a particular question in the future while they are working on questions. Specifically, within our well-trained model, for a student $s$ and a question $q$, we select the historical sequence of exercises attempted by the student before timestamp $t$. Then, for each moment before $t$ when the student has attempted other questions, we predict the probability of the student $s$ correctly answering question $q$. This process allows us to evaluate the student's mastery degree of question $q$ during the learning process. We visualize this mastery level, displaying information such as correctness, time intervals between questions, and whether they belong to the link multiset of $s$ and $q$, all depicted in figure \ref{Fig:case study ablation}. 

The boxes with different colors in the time interval line represent different time interval ranges between the current and next steps; the darker colored circles indicate that the attempted question belongs to the predicted link multiset.

\end{document}